 \documentclass[final,3p,times]{elsarticle}

\usepackage{amssymb}
\usepackage{graphicx}
\usepackage{xcolor}
\usepackage{enumitem}
\usepackage{float}

\begin{document}

\begin{frontmatter}



\title{SIDU-TXT: An XAI Algorithm for NLP with a Holistic Assessment Approach}


\author[label1]{Mohammad N.S. Jahromi\corref{cor1}%
\fnref{fn1}}
\ead{mosa@create.aau.dk}
\cortext[cor1]{Corresponding author}
\author[label1]{Satya. M. Muddamsetty\fnref{fn1}}
\ead{smmu@create.aau.dk}

\author[label2]{Asta Sofie Stage Jarlner}
\ead{assj@jur.ku.dk}

\author[label2]{Anna Murphy Høgenhaug}
\ead{ahh@jur.ku.dk}

\author[label2]{Thomas Gammeltoft-Hansen}
\ead{tgh@jur.ku.dk}

\author[label1]{Thomas B. Moeslund}
\ead{tbm@create.aau.dk}

\fntext[fn1]{These authors contributed equally to this work.}

\affiliation[label1]{organization={Visual Analysis and Perception Laboratory~(VAP),~Aalborg University},
            postcode={9000 Aalborg},
             country={Denmark}}

\affiliation[label2]{organization={Faculty of Law, Center of Excellence for Global Mobility Law.~University of Copenhagen},
            postcode={2300 København S}, 
            country={Denmark}}
\date{}

\begin{abstract}
Explainable AI (XAI) aids in deciphering 'black-box' models. While several methods have been proposed and evaluated primarily in the image domain, the exploration of explainability in the text domain remains a growing research area. In this paper, we delve into the applicability of XAI methods for the text domain. In this context, the 'Similarity Difference and Uniqueness' (SIDU) XAI method, recognized for its superior capability in localizing entire salient regions in image-based classification is extended to textual data. The extended method, SIDU-TXT, utilizes feature activation maps from 'black-box' models to generate heatmaps at a granular, word-based level, thereby providing explanations that highlight contextually significant textual elements crucial for model predictions. Given the absence of a unified standard for assessing XAI methods, this study applies a holistic three-tiered comprehensive evaluation framework—Functionally-Grounded, Human-Grounded and Application-Grounded—to assess the effectiveness of the proposed SIDU-TXT across various experiments. We find that, in sentiment analysis task of a movie review dataset, SIDU-TXT excels in both functionally and human-grounded evaluations, demonstrating superior performance through quantitative and qualitative analyses compared to benchmarks like Grad-CAM and LIME. In the application-grounded evaluation within the sensitive and complex legal domain of asylum decision-making, SIDU-TXT and Grad-CAM demonstrate comparable performances, each with its own set of strengths and weaknesses. However, both methods fall short of entirely fulfilling the sophisticated criteria of expert expectations, highlighting the imperative need for additional research in XAI methods suitable for such domains.
\end{abstract}








\end{frontmatter}


\section{Introduction}
Deep learning has substantially advanced Natural Language Processing (NLP), enabling significant progress in tasks ranging from sentiment analysis \cite{zhang2019integrating} to machine translation and and chatbot development  \cite{bahdanau2014neural}. Despite these advances, a critical challenge persists—the 'black-box' nature of such models. While the field of NLP continues to evolve, the deployment of these models in products and services of critical importance necessitates adherence to regulatory frameworks such as the GDPR and the forthcoming AI Act. This underscores the need for Explainable AI (XAI), ensuring that these advanced systems are transparent, accountable, and in line with public expectations \cite{bibal2022attention,mcgregor2023explainable}. The subjective nature of explanation quality, coupled with the intricacies of evaluations, further complicates the pursuit of transparency. XAI, especially as applied to text, lacks a standardized framework, leading to diverse interpretations of what constitutes understandability, comprehensibility, and interpretability \cite{cerutti2017interpretability}. These terms, often used interchangeably, reflect the need for clarity and uniformity in the evaluation of XAI methodologies. The current landscape of XAI evaluation is marked by a patchwork of approaches \cite{camburu2018snli}, each with its own merits and limitations, underscoring the necessity of developing more robust and tailored approaches that resonate with human intuition and expert judgment. This is particularly important in high-risk applications, e.g., those in the legal and medical domains, where the consequences of model outcomes can be significant. In parallel, for such applications, gaining trust from domain experts through assessing and verifying the decision quality of 'black-box' models is inevitable~\cite{doshi2017towards}.

In the realm of XAI research, one popular approach to make these models understandable to human end-users is to create a separate model that identifies the importance of given input features to the model being explained. This category of methods is referred to as 'post-hoc' explainability \cite{danilevsky2020survey}. The model agnostic LIME \cite{ribeiro2016model}  and class activation based GRAD-CAM \cite{selvaraju2017grad} are well-known examples of this explanation class that are widely used for image input. In spirit of these methods, the 'Similarity Difference and Uniqueness' (SIDU) method~\cite{muddamsetty2022visual}, a class activation-based explanation approach, was developed to effectively localize all the important features in an image responsible for the prediction.  Comprehensive quantitative and expert-driven qualitative analyses have demonstrated that this method not only outperforms both LIME and GRAD-CAM but also invokes greater trust from human users. \cite{muddamsetty2022visual}.

While these methods have set a standard in the visual domain, the text domain presents its own unique set of challenges and opportunities for explainability. Text data, with its complex syntactic and added semantic layers, requires an approach that can unravel the contributions of individual words and contextual relationships to a model's output. Inspired by the success of XAI methods in the visual domain, this work extends the class activation-based explanation algorithm 'SIDU' to the text domain, which we refer to as 'SIDU-TXT'. This adaptation aims to provide interpretability comparable to image-based tasks by generating 'heatmap' representations at a granular level for a given input text sequence. These 'heatmaps' emphasize the significance of individual words or contexts, effectively highlighting the textual elements that are crucial for the model's predictions. In transitioning from the visual to the textual realms, our approach aligns with existing scholarly observations that the main challenge in XAI for NLP lies in crafting explanations that are intuitively comprehensible to humans in the context of text analysis~\cite{hajiyan2022comparative,jain2019attention}. Thus, we have modified the SIDU algorithm to fine-tune the interpretability for NLP tasks. The key refinement in SIDU-TXT is the introduction of feature importance weights that represent similarity difference, and uniqueness scores, which are used to identify and select a subset of feature masks that carry the most semantic information for the given text features.

Moving beyond specific XAI tools, we confront the broader question of evaluating such methods in the NLP landscape. An inaccurate or low-fidelity explanation model can significantly limit trust in the explanation and, by extension, in the 'black-box' model it aims to explain. Currently, there is no consensus on how best to assess XAI techniques. However, an effective explanation should be \textit{faithful},  \textit{justifiable}, and \textit{comprehensible }to humans,  as well as invoke further \textit{trust} in expert users. Recognizing the need for a comprehensive evaluation of these aspects in NLP, we advocate for a holistic approach to the XAI assessment, encompassing Functionally-Grounded, Human-Grounded, and Application-Grounded framework~\cite{doshi2018considerations}, as illustrated in Fig~\ref{fig:evaluation_framework}. Concretely, we use sentiment analysis as a typical NLP task to evaluate SIDU-TXT through Functionally-Grounded and Human-Grounded approaches. For Application-Grounded assessment, we adopt the more complex task of asylum case classification in the sensitive legal domain. This comprehensive strategy ensures a thorough examination of the model's performance across various dimensions, from quantitative faithfulness to qualitative human expert involvement, ultimately fostering greater alignment with human intuition and enhancing the trustworthiness of explanations provided by SIDU-TXT. Our main contributions can be summarized as follows:
\vspace{0.3cm}
\begin{enumerate}
    \item We present SIDU-TXT, a powerful class activation-based XAI method optimized for NLP, which elucidates the decision-making process of 'black-box' models. By leveraging the top feature activation masks extracted from the models, along with their corresponding importance weights, SIDU-TXT discloses the most influential features driving the model's predictions. This method provides granular and insightful explanations at the word and context levels, significantly enhancing transparency and interpretability.
        
    \item A comprehensive quantitative assessment on the IMDB movie review dataset~\cite{maas2011learning} for sentiment analysis is conducted to assess the \textit{faithfulness/fidelity}. Our approach shows a significant improvement in both insertion and deletion metrics compared to established XAI methods such as LIME and GRAD-CAM, indicating greater precision in capturing and retaining critical textual features.

    \item We employ a Human-Grounded evaluation to analyze the \textit{justifiability} and \textit{comprehensibility} of XAI methods This process involved using human-provided explanations to assess the models at both token and sentence levels. The findings demonstrate that SIDU-TXT  excels in these criteria, achieving a higher Jaccard similarity score in justifiability and superior mean precision, recall, and F1-score in comprehensibility.
    \item For the first time, we apply an XAI model to complex and lengthy texts from the legal domain, with a particular focus on asylum decision-making where \textit{trust} in algorithmic judgment is paramount. This challenging application emphasizes the adaptability of SIDU-TXT to high-stakes domains where interpretability is crucial. Our study demonstrates that involving domain experts in the evaluation process significantly bolsters the trust and reliability of an AI system's decision-making capabilities.

\end{enumerate}

\section{Related Work}\label{sec1}
This section explores two intertwined dimensions in the realm of explainable AI (XAI) for the text domain. The first dimension, as covered in Subsection 2.1, presents an overview of various XAI methods that have been developed to provide clarity on how models process and analyze text data. The second dimension, outlined in Subsection 2.2, delves into the diverse strategies and challenges associated with evaluating XAI methods.

\subsection{XAI Methods for Text}

In the domain of XAI, various methods have been developed to demystify the decision-making processes of deep learning models in text classification, as highlighted in several surveys and studies \cite{danilevsky2020survey,li2016understanding,liu2019towards}. Among these, post-hoc or feature attribution methods stand out as particularly popular. They seek to pinpoint the most influential components of a given input text using techniques such as input perturbation \cite{ribeiro2016model}, gradient analysis, and relevance propagation \cite{arras2017explaining}. A recent survey specifically targeting NLP-focused XAI categorizes contemporary literature, mainly encompassing local post-hoc, local self-explaining, global post-hoc, and global self-explaining methods \cite{danilevsky2020survey}. These 'black-box' explanation methods comprise a variety of techniques, including saliency mask generation methods such as LIME or Grad-CAM, as well as architecture-specific methods like attention mechanisms \cite{guidotti2018survey}. While these approaches are often employed in the context of image data, their adaptability for text models has also been demonstrated, aiming to enhance transparency in model predictions. Well-established methods such as LIME and Grad-CAM, traditionally associated with Convolution Neural Networks (CNNs) for text classification and examined in earlier survey papers \cite{luo2021local}, are pertinent to our discussion. Our SIDU-TXT method, which generates saliency-based heatmaps, will employ these established methods as a benchmark for comparison.

\subsection{Evaluation of XAI Methods}
Evaluating the effectiveness of XAI methods involves a spectrum of techniques, each with unique benefits and limitations. Initial approaches often focus on faithfulness metrics (Functionally-Grounded), which may include strategies like word deletion to observe the impact of token (word) removal on prediction confidence \cite{doshi2017towards,nguyen2018comparing,du2019attribution}. These methods provide a quantitative means to assess explanation quality but may not fully capture the root cause of a model's decision, leaving the critical ‘why’ question only partially answered \cite{cambria2023survey}. Other quantitative methods involve using relevance scores to create document vectors, subsequently evaluated with traditional machine learning techniques \cite{xiong2018looking}. However, while feature importance metrics are widely utilized, their limitations are especially pronounced in the context of NLP in demonstrating the underlying rationale behind a linguistic phenomenon \cite{nauta2023anecdotal}. This gap highlights that the evaluation of XAI is still evolving and that there is a need for more nuanced approaches that can more clearly explain the model's reasoning in a way that aligns with human understanding.
Further methods, such as hybrid documents and morphosyntactic agreements, attempt to verify the accuracy of explanations without direct human involvement  \cite{poerner2018evaluating}. Yet, the ultimate goal of 'explanation evaluation' is not simply to select the most likely or accurate explanation from the model's perspective, but rather to identify explanations that possess pragmatic value—those that are the most useful and relevant to end-users, especially in sensitive real-world applications like the legal domain, where the implications of AI's decisions are critical.
In contrast, Human-Grounded evaluations bring in expert-derived relevance benchmarks and involve tasks where humans infer classifier outputs based on highlighted words, adding a layer of a qualitative dimension to the assessment process \cite{doshi2017towards, nguyen2018comparing,mohseni2018human}. We will embrace a dual approach, integrating both quantitative Functional-Grounded evaluation and qualitative human-expert judgment to validate the effectiveness of the SIDU-TXT method compared to algorithms like LIME and GRAD-CAM. This holistic evaluation strategy not only adheres to conventional metrics but also seeks to capture the pragmatic aspects of explanations in real-world contexts.

\section{SIDU-TXT}

The advent of deep neural networks in NLP has brought about transformative changes, yet the interpretability of these models remains a significant challenge. Here, we introduce SIDU-TXT, a novel approach to XAI that builds upon our previous success with SIDU in the visual domain, tailored to meet the intricate demands of text interpretation. SIDU-TXT is designed to illuminate the decision-making process of NLP models, providing granular insights into the textual elements that drive predictions. When adapting SIDU to text input, our focus shifts to the nuanced world of text classification. We aim at a method that identifies and emphasizes the subtle yet critical language elements—ranging from individual words to phrases, or N-grams—that are pivotal in text-based decision-making of a text classification model. This capability sets SIDU-TXT from current state-of-the-art methods, which often struggle to highlight the specific sequence of words or phrases that provide essential contextual information when applied to the domain of textual data.

The core concept of SIDU-TXT centers on deriving feature text masks from the last convolution layers of a CNN model. It then utilizes the most significant $K$ masks to compute two key metrics: similarity differences (SID) and uniqueness (U). SID assesses the importance of text features by measuring the change in the model's output when a feature is altered, indicating its influence on the prediction. Uniqueness evaluates the distinctiveness of a feature's contribution compared to all other features, with more unique features receiving higher importance. The final explanation, a heatmap, emerges from a weighted combination of these feature masks, offering a visual representation of the text's influential elements as discerned by the model. 

SIDU-TXT begins by extracting feature text masks from the final convolutional layer of a deep CNN model, denoted as  $\textit{'F'}$. Unlike in image processing, where convolution layers capture spatial visual features, in NLP, these layers are interpreted as capturing syntactic and semantic patterns in textual data. This distinction is important, as the nature of text data demands a different approach to feature extraction and interpretation. The layers, denoted by dimensions $'n \times N'$, play a central role in pinpointing textual features that are relevant for a specific class 'c', formalized as $f^c_{i} = \left[f^c_1, f^c_2, \ldots, f^c_N\right]$. For each textual feature activation $f^c_i$, we derive a
corresponding binary mask $B^c_i$ through a thresholding operation:
\begin{equation}
B^c_{i=1,...,N} = H(f^c_{i=1,...,N} > \tau)
\end{equation}

Where, $B^c_i$ denotes the binary mask for the ith feature, and $\tau$ represents the threshold value,
and the expression $f^c_i > \tau$ is a condition that a function $H$ uses to set each element of $B^c_i$ to '1' if $f^c_i$
exceeds $\tau$, and '0' otherwise. This process effectively filters the activation maps, retaining only those features that surpass the threshold, thereby highlighting the most influential text components for class 'c'. Note that, $\tau$ is a hyperparameter set experimentally to balance between interpretability and noise reduction in heatmap visualization.

\begin{figure}[htbp]
  \hspace*{-1.5cm} 
\includegraphics[width=\linewidth, height=0.56\linewidth]{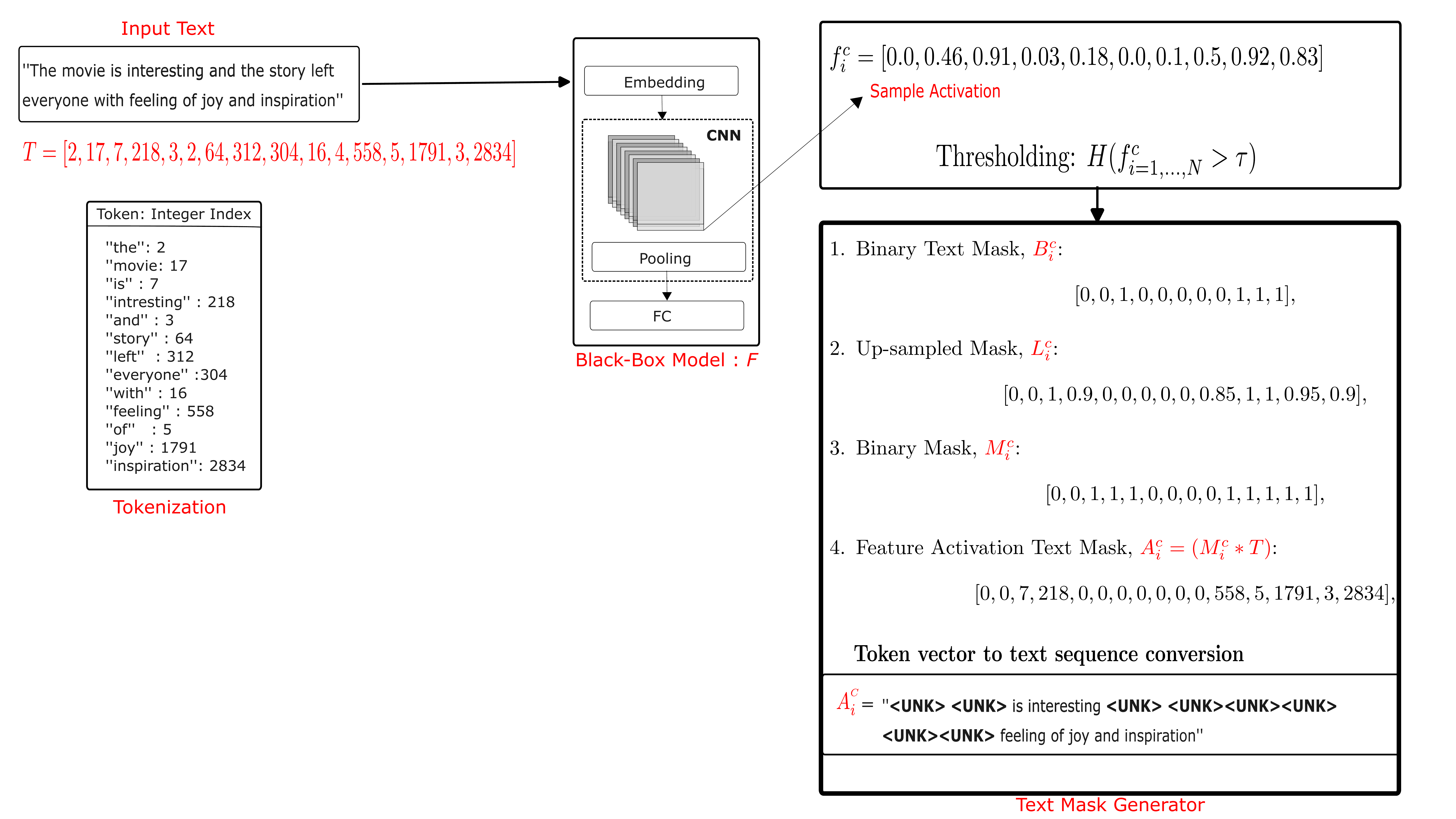}
\caption{The block diagram depicts the process of generating a feature activation text mask, serving as input for the SIDU-TXT XAI method. Within the diagram, $\langle$unk$\rangle$ symbolizes an unknown word, typically employed in NLP to indicate a missing word in a sentence. This token is mathematically assigned a value of 0.}
  \label{fig:mask}
\end{figure}

The binary text mask \( B^{c}_{i} \) is then upscaled to match the input text's sequence length \( T \) using bilinear interpolation, ensuring that the feature activations correspond to the actual text sequence. The resulting up-sampled mask, denoted as \( L^{c}_{i} \), contains values within the range \([0,1]\). This up-sampled mask is further converted back into a binary mask, represented as \( M^{c}_{i} \). To effectively distill the most influential text features as learned by the model via its convolution layer for a given text input, the binary mask \( M^{c}_{i} \) is combined with the input text token vector \( T \) through point-wise multiplication. This generates the feature activation text mask \( A^{c}_{i} \), defined as:
\begin{equation} 
  { A^{c}_{i} = T \odot M^{c}_{i},}
\end{equation}

where, $T$ is an input text, and $M^{c}_{i}$ is an up-sampled binary mask. The procedure for feature activation's text mask generation is illustrated in Figure~\ref{fig:mask}.

After generating the upscaled feature text activation masks, the prediction score vectors $P^{c}_{i}$ and $P^{c}_{org}$ are computed for all feature activation text masks $M^{c}_{i}$, and the original input text $T$ using the 'black-box' model. We then proceed to determine the similarity difference between each prediction score  $P^{c}_{i}$ associated with an input feature activation text mask and the prediction score $P^{c}_{org}$ of the original input text $T$.  The similarity difference  $SID^{c}_{i}$ between these two vectors signifies the relevance of the feature activation text mask concerning the original input text $T$ and it is defined by utilizing the Gaussian kernel as:

\begin{equation}\label{equnic}
    SID^{c}_{i} = \exp\left(\frac{-\|P^{c}_{org}-P^{c}_{i}\|}{2\sigma^2}\right)
\end{equation}

where $\sigma$ represents the spread of the kernel, effectively controlling the sensitivity of the similarity measure to differences between the prediction scores. This step aims to minimize the influence of less significant textual elements with low weights while highlighting those textual elements genuinely responsible for the model's predictions. Following the calculation of the similarity difference measure, we also compute a uniqueness measure $U^{c}_{i}$  to identify distinct textual feature masks among all the prediction score vectors of the feature activation text masks.  The uniqueness measure is defined as:

\begin{equation}\label{equsd}
    U^{c}_{i} = \sum_{j=1}^{N}\|P^{c}_{i}-P^{c}_{j}\|,{~~~~~~~i= 1,2,..., N}
\end{equation}
The SIDU weights are computed as the product of the similarity differences and uniqueness measures and is given by
\begin{equation}
    W^{c}_{i}= SID^{c}_{i}\cdot U^{c}_{i},
\end{equation}

where $SID^{c}_{i}$ is the similarity difference and $U^{c}_{i}$ is the uniqueness measure given in Eq. (\ref{equnic}) and Eq.  (\ref{equsd}), respectively. Upon computing the weights, we then introduce a step tailored to the textual domain: selecting the most informative masks based on their weights $ W^{c}_{i}$.
This process entails arranging the weights in descending order and focusing on the top-ranking masks. For instance, if we opt for the top $K$ weights, only the top $K$ masks are considered, where each mask can represent different aspects of language use, such as semantic nuances or syntactic structures. The motivation for this selective approach stems from our careful examination of all generated feature activation masks. Masks with higher weights are found to signify features rich in semantic information, essential for the model’s decision-making process. On the other hand, masks with lower weights tend to contain less important information. This step contrasts with image processing methods, where the significance of spatial and visual features follows a different pattern of relevance.

\begin{figure}[htbp]
  \centering
    \includegraphics[width=\linewidth, height=0.5\linewidth]{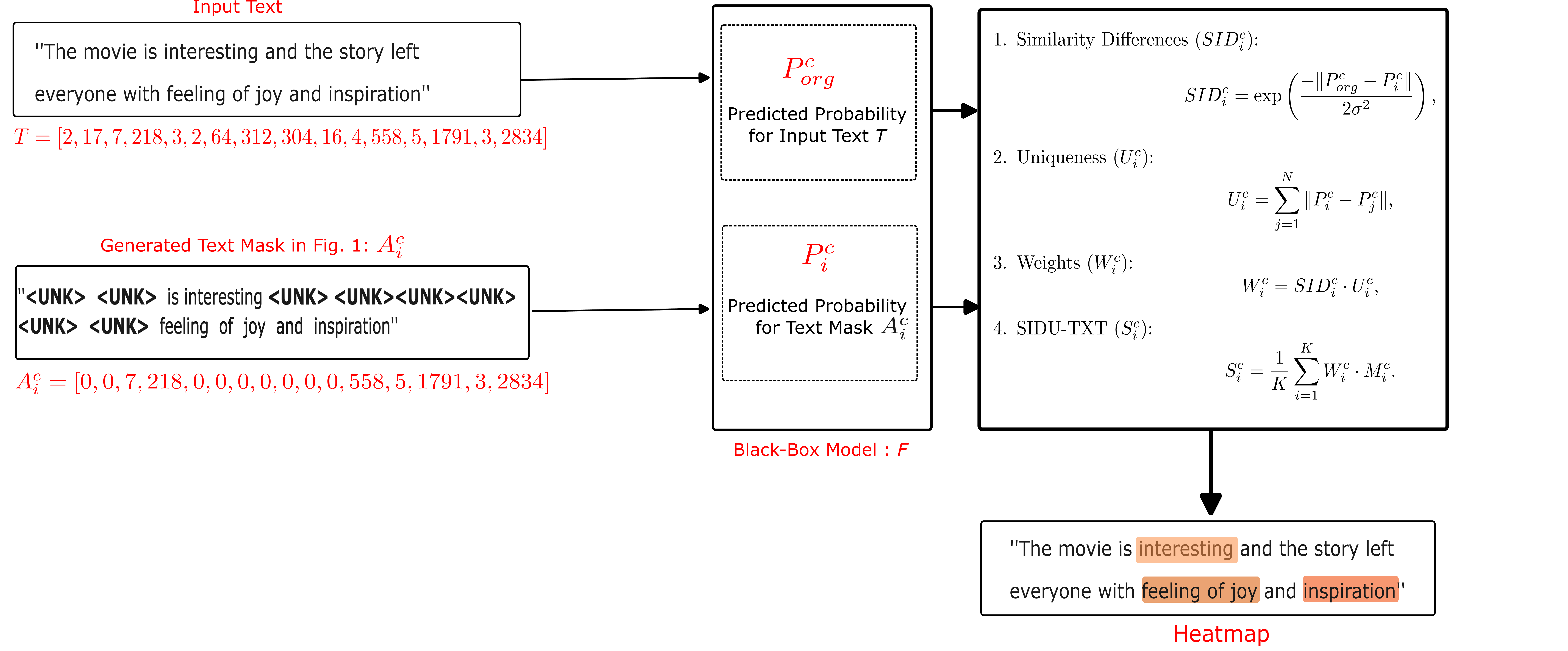}
  \caption{Block diagram illustrates the procedure of generating the explanations for the text data using SIDU-TXT.}
  \label{fig:visual_exp}
\end{figure}

Ultimately, the final explanatory output of SIDU-TXT, $S^{c}_{i}$ ,is derived by calculating a weighted sum of the feature activation text masks with the highest weights:
\vspace{-.1cm}
\begin{equation} \label{eq: visual_exp}
    S^{c}_{i} = \frac{1}{K}\sum_{i=1}^{K} W^{c}_{i} \cdot M^{c}_{i},
\end{equation}
where $K$ represents the number of top masks used for text explanation, a hyperparameter that primarily influences the the granularity of the explanation generated by SIDU-TXT. 
An overview of SIDU-TXT is summarized in Figure~\ref{fig:visual_exp}.

\section{ Assessing XAI Methods } \label{sec:METHODS}
\begin{figure}[htb]
  \centering
  \includegraphics[width=\linewidth,height=4.5cm]{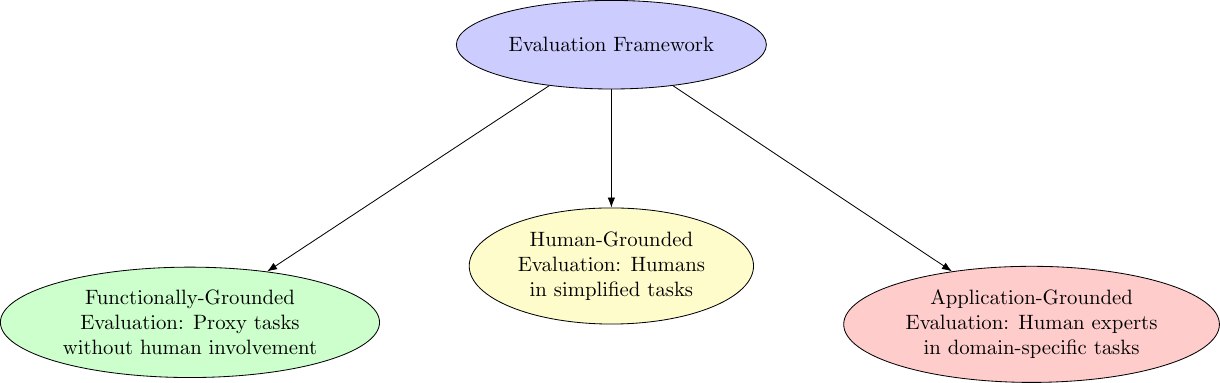} 
  \vspace{-0.2cm}
  \caption{Three aspects of empirical assessment within the  XAI Evaluation Framework for text.}
  \label{fig:evaluation_framework}
\end{figure}

This section lays the groundwork for a comprehensive evaluation framework, drawing from interdisciplinary insights into what constitutes an effective explanation. Building upon the challenges and opportunities discussed in our related work, we articulate a set of criteria any XAI method should strive to meet: \textit{fidelity/faithful} to the model’s decision process, \textit{justifiability}, \textit{comprehensibility} in the eyes of users, and \textit{trustworthy} by the expert audience. These criteria, informed by the selected taxonomy~\cite{doshi2017towards}, will guide our empirical assessment in the upcoming experiments section. 
The approach, as illustrated in Fig~\ref{fig:evaluation_framework}, provides a versatile framework for XAI NLP evaluation and allows for the domain-specific customization of experiments, which makes it suitable for our objective.
We will explore these criteria in the subsequent subsections: \ref{sec:41} Functionally-Grounded Evaluations, which focus on proxy tasks without human involvement; \ref{sec:42} Human-Grounded Evaluation, involving real humans in simplified tasks; and \ref{sec:43} Application-Grounded Evaluation, engaging human experts in domain-specific tasks.

\subsection{Functionally-Grounded Evaluations: No Humans, Proxy tasks}\label{sec:41}

Functionally-Grounded Evaluation focuses on assessing XAI methods without the direct involvement of human participants. This approach, while efficient in bypassing the complexities of human-subject experiments, challenges us to identify appropriate proxies that accurately reflect decision-making features. By utilizing formal definitions of explanations, we define a proxy function to evaluate the \textit{faithfulness} of XAI methods. To assess the faithfulness of XAI explanations, we employ Insertion and Deletion metrics, adapted from image processing applications to the text domain. These metrics gauge the impact of highlighted text on model predictions, offering insights into the explanations' influence on classification scores. A significant change in probability scores upon inserting or deleting words indicates the degree of faithfulness of the explanation method.

The practical implementation of this evaluation involves training a text classification model on a specified text corpus. Upon classification of input text, the chosen XAI method is applied to elucidate the model’s predictions, focusing on the most significant text fragments. This procedure highlights our approach to validate the faithfulness of explanations generated by XAI methods.

\subsection{Human-Grounded Evaluation:  Humans in Simplified Tasks}\label{sec:42}

In the realm of Human-Grounded evaluation, the focus shifts to conducting practical, human-centric experiments that mirror the fundamental aspects of the intended application. This methodology is especially beneficial when direct experiments with the target audience are impractical. By involving non-expert humans in these evaluations, we gain the advantage of a potential participant pool and reduced costs, as the need for specialized domain experts is eliminated. Key to this evaluation is the task of assessing the \textit{justifiability} and \textit{comprehensibilty} of explanations provided by XAI methods. In these tasks, human reasoning and intuition serve as benchmarks for evaluating the explanations generated by XAI systems. Essentially, we seek to determine if these explanations resonate with the human understanding of the decision-making process.
To this end, we design tasks that involve systematic data collection to measure the alignment between human annotations and XAI explanations. Participants engage in activities such as identifying key elements—tokens and sentences—that they perceive as influential in the decision-making context of the model. The effectiveness of XAI methods is then evaluated based on their ability to mirror these human-selected elements. This approach not only tests the justifiability of the XAI methods but also enriches our understanding of how closely machine-generated explanations align with human thought processes.

\subsection{Application-Grounded Evaluation: Human experts in Domain-specific Tasks}\label{sec:43}
In real-world scenarios, the effectiveness of XAI is often measured by its contribution to crucial objectives like enhancing safety or promoting trust. This evaluation approach, particularly relevant in contexts where specific applications are targeted, involves detailed analyses by human experts within real-world scenarios. It is especially valuable in fields such as medical diagnostics or legal judgments, where the impact of model decisions can be substantial.This section aims to evaluate the \textit{trustworthiness} of XAI methods through the lens of domain-specific experts. It seeks to determine if the explanations provided by these methods can help experts grasp the decision-making process of the underlying models. Such understanding is vital in high-stakes areas where model decisions carry significant consequences.

To conduct this evaluation, we utilize models trained on domain-specific tasks and apply different XAI methods to generate explanations for the model's predictions. These explanations, along with the predicted outcomes and their associated probabilities, are presented to domain experts for assessment. The experts are charged with identifying which XAI method best clarifies the model’s reasoning in a manner consistent with their expert judgment. An XAI method favored by the experts would indicate its effectiveness in enhancing understanding and building trust in the 'black-box' model.

\section{Experiments and Results}
This section presents the experimental study of our SIDU-TXT method, employing the threefold XAI assessment framework detailed in Section~\ref{sec:METHODS}, aimed at evaluating the model's interpretative capabilities. Initially, we describe the datasets utilized for our analysis, highlighting their specific features and how they align with our research goals. We then describe the training process of the AI models, tailored for two distinct applications: sentiment analysis of the IMDB movie reviews and asylum decision prediction. The subsequent subsections delve into the results derived from our Functionally-Grounded, Human-Grounded, and Application- Grounded evaluations. These results offer further insights into the performance and interpretability of the XAI methods employed, highlighting their practical efficacy in practical implementations.

\subsection{Datasets} \label{dat}

 \begin{enumerate}
     \item The \textbf{IMDB} (Internet Movie Database) dataset is a widely-used resource for sentiment analysis in NLP~\cite{maas2011learning}. It consists of a large collection of movie reviews, each labeled as either 'positive' or 'negative', making it an ideal benchmark for binary sentiment classification tasks. The dataset typically comprises 50,000 reviews, evenly split into training and testing sets. Each review is a variable-length text covering a diverse range of movies, offering rich linguistic content for analysis. The reviews in the IMDB dataset are user-generated, reflecting a variety of writing styles and opinions, which adds to the complexity and realism of sentiment analysis tasks. This dataset provides a comprehensive platform for developing and evaluating models aimed at understanding and classifying human emotions and opinions expressed in written form.
    
     \item The\textbf{ Danish asylum decisions dataset}, provided by the Refugee Appeals Board, consists of 14,987 asylum case files from January 1995 to January 2021. It reflects the distinct nature of the Danish two-tiered asylum process. Each file offers detailed information about asylum seekers, including personal backgrounds and specifics of their asylum claims~\cite{unhcr2023nordic}. The dataset is enriched with narrative accounts, legal reasoning, and various supplementary documents, making it a rich resource for NLP-based legal analysis. The dataset comprises cases with diverse outcomes: 15\% are granted, 83\% rejected and 2\% are sentback cases (pending cases). These figures underline the dataset's diversity and complexity, offering a realistic representation of asylum decision-making in Denmark. This comprehensive data collection serves as an effective tool for modeling and understanding the nuances of asylum adjudication using deep learning techniques in NLP.
     
 \end{enumerate}
\subsection{Training  a Model for Sentiment Analysis and Asylum Decision Prediction}
Before delving into the XAI analysis, a predictive AI model is needed. In our study, we leverage Convolutional Neural Networks (CNNs) to address two distinct text classification challenges: sentiment analysis of the IMDB movie reviews and decision prediction for Danish asylum cases. CNNs effectively process text by translating it into dense vector representations through an embedding layer, followed by the extraction of features using convolution layers to identify key patterns like n-grams. The extracted features are processed through fully connected layers, where softmax or sigmoid functions are used to determine the probability of each class, enabling precise class differentiation. This comprehensive process makes CNNs well-suited for our varied text classification tasks. In addition, the hyperparameter tuning strategy and hardware specification are briefly discussed.

For each task, we design a specific CNN model, tailored to the unique characteristics and requirements of the respective datasets. This involves customizing the architecture and training process to best fit the nature of the data and the objectives of the analysis. The following subsections detail the development, training, and evaluation of each model, highlighting the nuances that make each model suited for its specific task.

\subsubsection{IMDB Sentiment Analysis Model}\label{senmodel}
In the sentiment analysis of the IMDB movie review dataset, we develop a CNN model using TensorFlow~\cite{tensorflow2015}, optimized for distinguishing between positive and negative reviews. The model employs GloVe pre-trained embeddings~\cite{pennington2014glove} in its initial embedding layer with a dimensional of 300, leveraging these established word vectors to capture the semantic richness of the reviews. This layer is followed by a convolutional layer with 128 filters and a kernel size of 5, using \textit{'ReLU'} activation to extract text features relevant to sentiment analysis.
To enhance the model's generalizability and reduce over-fitting, we incorporate dropout layers with a rate of 0.2, following the convolutional and dense layers. The convolutional layer's output undergoes dimensionality reduction through a \textit{'GlobalAveragePooling1D'} layer, leading to a dense layer of 64 neurons with \textit{'ReLU'} activation. The final output layer comprises a single neuron with a \textit{'sigmoid'} activation function, making it suitable for binary classification. For our training and validation purposes, we utilize the pre-structured split of the IMDB dataset, which comprises 50,000 reviews divided equally into training and testing sets. From the training portion, we further allocate 15\% for validation, effectively utilizing the dataset for a comprehensive evaluation of our model. The training was conducted over 30 epochs with a batch size of 32. The model was compiled using \textit{binary cross-entropy} as the loss function and optimized with an \textit{'Adam optimizer'} set to a learning rate of 0.0002, and its performance is primarily evaluated based on accuracy. This architecture, combining GloVe embeddings with a carefully structured CNN, provides a robust framework for effective sentiment classification. The model is trained on the balanced dataset and achieved an overall accuracy, precision, recall, and F1-score of $90\%, ~89.577\%,~89.567\%$ and $89.566\% $ respectively, on the test dataset.


\subsubsection{ Asylum Case Decision Prediction Model}
In the context of asylum case decision prediction, the CNN model is specifically trained on the Danish asylum dataset, as described in Section ~\ref{dat}, to categorize case files as either 'Granted' or 'Rejected'. The dataset is partitioned into 80\% for training, 10\% for validation, and 10\% for testing. Pre-processing steps include the removal of punctuation, white spaces, and other insignificant symbols, as well as the lemmatization of text. Unlike typical preprocessing methods, stop words are retained due to their potential contextual relevance in legal texts.

The embedding layer, trained with the full vocabulary of the dataset, is set to produce 300-dimensional feature vectors.
This is followed by a convolutional layer with 128 filters, kernel size 4, and \textit{'ReLU' }activation, aimed at identifying key legal text features. To mitigate overfitting, dropout layers at a rate of 0.2 are employed, along with a \textit{GlobalAveragePooling1D} layer to reduce output dimensionality. The next stage consists of a dense layer with 64 neurons (activated by \textit{'ReLU'}), and the final output is produced by a dense layer with 2 neurons using\textit{ 'Softmax'} activation, suitable for categorical classification.

For optimization, we employ \textit{categorical cross-entropy} as the loss function and the \textit{Adam optimizer} with a
learning rate of 0.0002 and momentum of 0.9. The training spans over 30 epochs with a batch size of 8, balancing computational demands with model efficiency. An early stopping strategy is utilized based on validation loss, complemented by dropout layers for additional regularization. The asylum dataset has class imbalance, we introduced the class weights. Class weights alter the loss function directly by penalizing the classes with varying weights. The trained CNN-NLP predictive model achieved overall precision, recall and F1-score of $82.948\%,~85.188\% $~and  $83.184\%$ respectively. 


\subsubsection{Hyperparameter Optimization and Hardware Specifications}
In the development of our models for IMDB sentiment analysis and asylum case decision prediction, we utilize Keras Tuner for hyperparameter optimization. Keras Tuner employs the Hyperband algorithm~\cite{li2017hyperband}, an advanced optimization method that efficiently balances exploration and exploitation of the hyperparameter space. This approach allows us to effectively and rapidly identify optimal configurations, including layer structures and learning rates, within the constraints of our computational resources.

Both models are developed and trained on a system equipped with an Nvidia RTX 2080Ti GPU, providing 11GB of GPU memory. This hardware was crucial for managing the demanding computational needs of deep learning models, particularly in light of the extensive datasets and sophisticated model architectures we work with. The selection of this hardware was instrumental in achieving the best balance between high model performance and efficient training.

For the SIDU-TXT XAI analysis, we focus on two important hyperparameters: the threshold $\tau$ and the number of top masks $K$. We set the threshold at $\tau = 0.5$, for converting sample a value chosen experimentally to optimize the interpretability of text data within the heatmap. The selection of this threshold significantly affects the granularity of the interpretability, impacting how words are highlighted and their contextual meanings are represented. Unlike in image processing, where threshold variations often have minimal impact, in text analysis, this setting plays a key role in the quality of the heatmap visualization. Given this importance, our approach emphasizes balancing interpretability with noise reduction in heatmap visualization. This threshold setting, while optimal in our experiments, may be subject to further refinement for different NLP tasks to achieve more precise interpretability, respectively. In a similar mechanism, we select the top $K$ masks, setting $K$ to $10$, to generate an output visual explanation rich in contextual relevance. Masks with lower SIDU weights, offering limited context, are excluded from the final explanatory visualization, ensuring focus on the most significant textual features.%
\subsection{Results}
\subsubsection{Functionally-Grounded Evaluation Results}

This section presents the results of our Functionally-Grounded evaluation outlined in Section~\ref{sec:41}. Our focus here is on quantitatively assessing the faithfulness of XAI methods through a series of deletion and insertion experiments. These experiments aim to measure how the removal or addition of key text elements impacts model predictions. Such an approach provides a practical assessment of the XAI methods' fidelity to the underlying decision-making process of the model.
To conduct this evaluation, we utilize the model described in~\ref{senmodel}. The evaluation involves systematically deleting words (setting the word vector to zero) and observing the consequent effect on classification scores. Additionally, we inserted words in sequence, according to their relevance scores, to track their impact on model predictions. The faithfulness of the XAI methods is quantified using the Area Under Curve (AUC) metric for each text sample, with results derived from 100 randomly selected texts from the IMDB validation dataset.
Our findings, as represented in Table \ref{inse}, demonstrate the comparative performance of our proposed method against established XAI methods  LIME and GRAD-CAM. The results reveal that our method surpasses both LIME and GRAD-CAM in terms of the insertion and deletion metrics, with GRAD-CAM following in performance. LIME's approach, based on random perturbations, exhibited limitations in providing contextual relevance due to its inherent randomness in feature instance generation. This contrasts with SIDU-TXT, where feature text activation masks are derived from meaningful internal model activations, offering a more insightful and contextually relevant analysis.

\begin{table}[h]
\centering
\caption{Faithfulness evaluation of XAI methods measured in terms of mean AUC. }
\begin{tabular}{|l|c|c|} 
\hline
XAI METHOD & Insertion (mean~AUC) $\uparrow$ & Deletion (mean~AUC)  $\downarrow$ \\ \hline
SIDU-TXT       & \textbf{0.5513}    & \textbf{0.1537}   \\ \hline
GRAD-CAM   & 0.4308    & 0.2073   \\ \hline
LIME       & 0.4228    & 0.2431  \\ \hline
\end{tabular}
\label{inse}
\end{table}

\subsubsection{Human-Grounded Evaluation Resutls}
Building on the framework established in the Human-Grounded Evaluation section, we now present our quantitative findings. These results are derived from comparing the XAI-generated explanations with human annotations, a key aspect of our methodology to assess the justifiability of XAI methods. Employing lexical similarity measures, we quantitatively analyze the alignment between the explanations provided by XAI methods and the words highlighted by human annotators. This approach offers a concrete metric to evaluate the extent to which machine-generated explanations correspond with human reasoning.
Our study begins with the systematic collection of data to assess the alignment between human sentiment analysis and the visual heatmap provided by XAI methods. We collect $50$ short movie reviews from the IMDB database as the basis for this analysis. The aim is to compile a set of human annotations to act as a standard for comparison against the sentiment assessments generated by XAI. Four annotators, each reviewing the same set of 50 reviews,  participated in the study. They each read the reviews and classify them as either 'positive' or 'negative' based on their assessment of the overall sentiment. They are tasked with identifying up to 10 words or phrases that influence their sentimental decision. In addition, they highlight up to five sentences that they believe best exemplify the sentiment of the review. Figure~\ref{fig:4} shows an example of an annotation provided by one participant for a review from our dataset of 50 movie reviews. The final set of annotations is formed by creating a union of the inputs from all four participants. This union process involves compiling all unique words and sentences identified by any participant, ensuring no duplicates are included in the collective set. This aggregated data provides a robust benchmark for the subsequent evaluation of XAI methods, offering a substantial measure of the consistency between AI-generated interpretations and collective human judgment both at a token level and sentence level.

\begin{figure}[H]
\centering
\fbox{
\begin{minipage}{0.9\textwidth} 
\scriptsize
\begin{itemize}
    \item \textbf{Input Text} : ``Hobgoblins is a very cheap and badly done Gremlins rip-off. That's the best thing one can say about this stinkpile. Pretty much everyone in the cast was chosen for their looks and not their acting ability. It was very painful to watch. Avoid this one at all costs.''
    \item Label the Review:
    \begin{itemize}[label={}]
        \item Positive
        \item \textbf{Negative}
    \end{itemize}
    \item Characterizing Words: Based on your label judgment, list up to 10 words from the review that characterize its sentiment. These are words that you think significantly support your chosen label.
    \begin{itemize}[label={}]
        \item Cheap, badly done, stinkpile, painful, avoid
    \end{itemize}
    \item Significant Sentences or Contexts: Highlight or pick up to 5 sentences or contexts in the review that support your selected label. These are sentences or parts of sentences that you believe most clearly convey the sentiment of the review.
    \begin{itemize}[label={}]
        \item \colorbox{yellow}{``Hobgoblins is a very cheap and badly done Gremlins rip-off.''}
        \item \colorbox{yellow}{``That's the best thing one can say about this stinkpile.''}
        \item \colorbox{yellow}{``It was very painful to watch. Avoid this one at all costs.''}
    \end{itemize}
\end{itemize}
\end{minipage}
}
\vspace{-0.3cm}
\caption{Annotation sample from one participant for a movie review in the dataset.}
\label{fig:4}
\end{figure}

\begin{figure}[!h]
  \centering
  \includegraphics[width=0.85\linewidth,height=6cm]{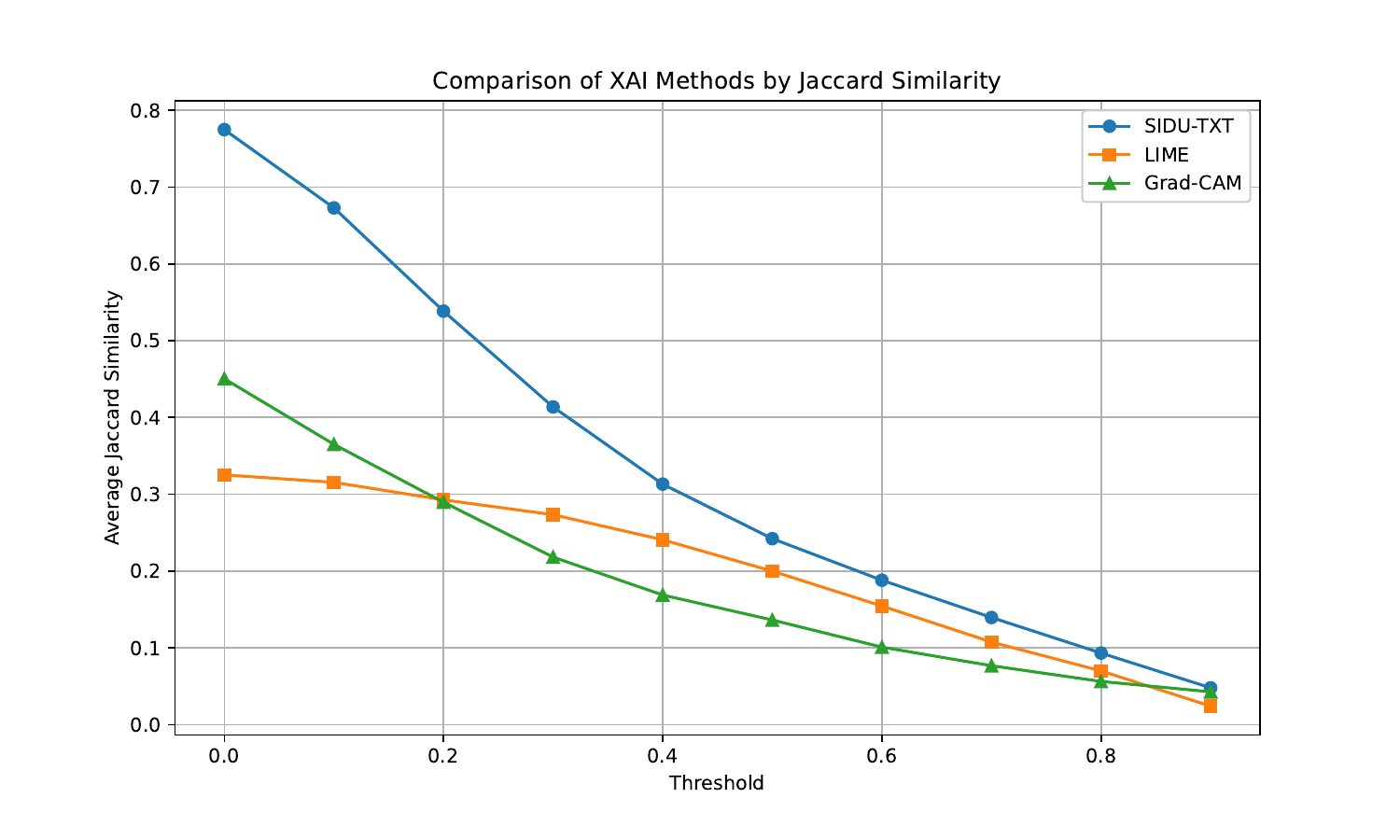}
  \vspace{-.65cm}
  \caption{Average Jaccard Similarity Scores between Human-Annotated token and XAI-Generated token score for $50$ Movie Reviews.}
  \label{fig:jac}
\end{figure}

In the next step, to quantitatively compare the words highlighted by XAI methods with human-collected annotations, one can utilize lexical similarity measures common in NLP applications. These measures assess the similarity between two text sets through the intersection of their word sets, irrespective of language. A lexical similarity score of 1 denotes complete overlap, indicating identical texts, whereas a score of 0 means there are no shared words. Such measures are useful in NLP for comparing word or phrase sets identified by different annotators or processes. The Jaccard similarity index and cosine similarity are among the popular methods used for such evaluation purposes~\cite{vijaymeena2016survey}. Our study employs the Jaccard similarity coefficient to facilitate direct quantitative comparison between influential words identified by human annotators and those highlighted by XAI techniques. We apply a threshold to the XAI scores to include only those words also selected by human annotators. This approach means that, for example, if humans identified up to 5 keywords, only the corresponding XAI scores exceeding the threshold are considered. At a threshold of 0, all words scored by the XAI method and identified by humans are included. As the threshold increases, stricter criteria are applied, and only words with an XAI score above the threshold are compared. For instance, at a threshold of 0.8, only words matching those identified by humans and with an XAI score above 0.8 are selected for comparison. The average Jaccard similarity results between 50 human-annotated movie reviews at the token level and XAI-generated token scores are presented in Figure \ref{fig:jac}.

As it can be seen in Figure~\ref{fig:jac}, the average Jaccard similarity decreases as the threshold increases for all three methods. This is because a smaller subset of words meets the criteria to be included in the comparison set between the XAI and the ground truth-- reducing the likelihood of overlap with the human-annotated set. At lower thresholds, SIDU-TXT exhibits the highest average Jaccard similarity, suggesting that when considering a broader set of words, SIDU-TXT aligns more closely with human judgment compared to the other methods. LIME maintains a moderate level of similarity across all thresholds, suggesting a consistent but less pronounced alignment with human annotations. On the other hand, while Grad-CAM shows steady performance at different threshold levels, it might not detect as wide a range of important words as SIDU-TXT. This highlights SIDU-TXT's sensitivity to a broader range of influential tokens, which is indicative of its superior alignment with human judgment in the sentiment analysis tasks we evaluated. The observed decline in similarity for all methods as the threshold increases also raises important considerations about the selection of a threshold for practical applications. A lower threshold may capture a larger, potentially more comprehensive set of influential words but at the risk of including less relevant tokens. Conversely, a higher threshold may result in a more focused set of words but could miss out on tokens that humans find influential. 
In conclusion, SIDU-TXT, particularly at a lower threshold, emerges as a strong candidate for scenarios where a comprehensive extraction of influential words is paramount.

Given the complexities discussed in XAI evaluation within the literature, which we have also addressed in the Related Work section, where factors that may be faithful to the model's rationale may not always be intuitive to human reasoning, in addition to our quantitative investigations, we further employ a qualitative study in our case of sentiment analysis. This involves examining the heatmaps generated by SIDU-TXT, LIME, and Grad-CAM, assessing them qualitatively against the annotations we collected. Such a comparison allows us to observe the degree of agreement between the AI-generated interpretations and the human perspective.

LIME's heatmap is produced through a process that randomly samples the vicinity of the input data and generates perturbations. This method assesses the model's behavior with these variations to identify both positive and negative contributions to the final prediction. By nature, LIME's technique for text analysis involves a duality: it highlights contributions that increase (in orange) or decrease (in blue) the model's confidence in its decision. The intensity of the color shades on the highlighted words in the heatmap illustrates the strength of each word's contribution, whether positive or negative, to the model's prediction. 
Grad-CAM generates its heatmap by applying the gradient-based localization technique to the last convolutional layer of the neural network. This process involves backpropagating the signal from the output layer to the text input level, thus producing a visualization that highlights words according to their impact on the model's decision-making process. SIDU-TXT, like Grad-CAM, utilizes the last convolutional layer of the given model. However, instead of tracing gradients, SIDU-TXT employs a Similarity and Uniqueness metric on a selected feature mask, as detailed in Section \ref{sec:METHODS}, to generate the heatmap of highlighted words. For both Grad-CAM and SIDU-TXT methods, shades of orange are utilized within the heatmaps to denote the strength of the scores assigned to words in each method, respectively. These shades correspond to a spectrum of ten intervals that represent the XAI methods' scoring range. The more intense the color shade, the closer the score is to 1, indicating a higher influence of the word on the model's prediction, and conversely, a paler color suggests a lower influence.
Table~\ref{tab:negative} and Table~\ref{tab:po} illustrate the qualitative comparison of XAI methods with respect to human token-level annotations for two samples (negative and positive) in sentiment analysis. To ensure consistency with human annotators who were instructed to identify up to $10$ influential words, we adapt our analysis to focus on the same number of words across all XAI methods. This offers a more standardized basis for comparison and maintains alignment with the human-provided annotations.

For our analysis, here we consider the  ’Negative sentiment' example and the corresponding heatmaps generated by the XAI methods as presented in Table \ref{tab:negative}.
\setlength{\tabcolsep}{1pt} 
\renewcommand{\arraystretch}{1.0} 
\begin{table}[!ht]
\centering
\caption{Comparative heatmaps of the top 10 words from 'Negative sentiment' IMDB dataset reviews. Yellow in Human Annotation denotes equal importance across words. LIME shows negative significance in varying shades of blue and positive in orange, while deeper shades in GRAD-CAM and SIDU-TXT indicate words of greater importance for the XAI methods.}
\begin{tabular}{|c|c|}
\hline
Method & Heatmaps \\
\hline
\centering\raisebox{5.5ex}{Human Annotation} & \fbox{\includegraphics[width=0.75\linewidth]{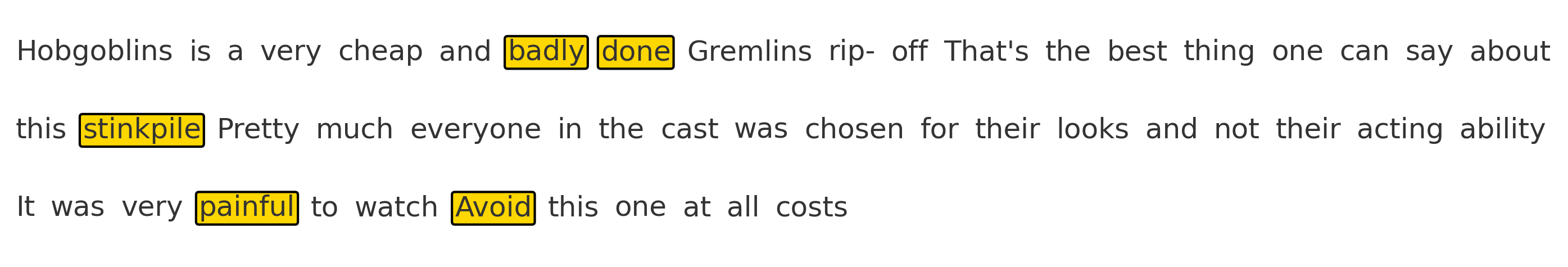}} \\
\hline
\centering\raisebox{12ex}{LIME} & \fbox{\includegraphics[width=0.75\linewidth,height=4.6cm]{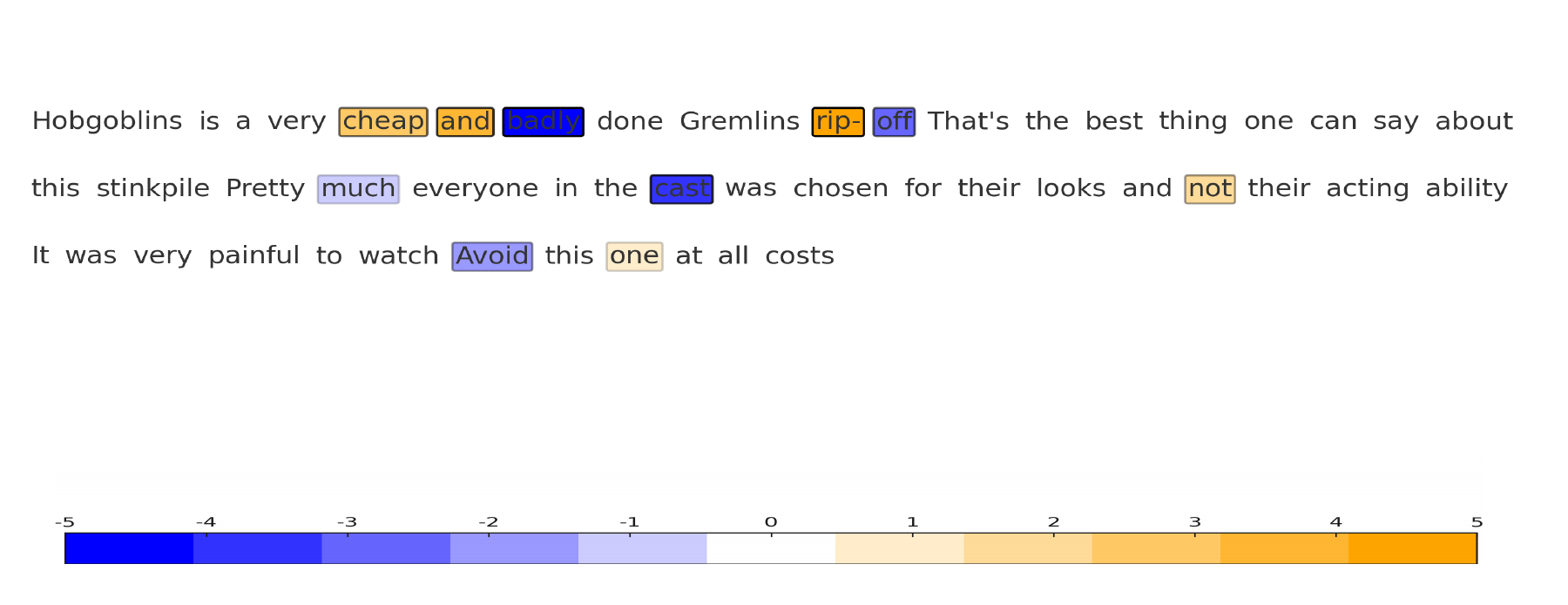}} \\
\hline
\centering\raisebox{12ex}{Grad-CAM} & \fbox{\includegraphics[width=0.75\linewidth,height=4.6cm]{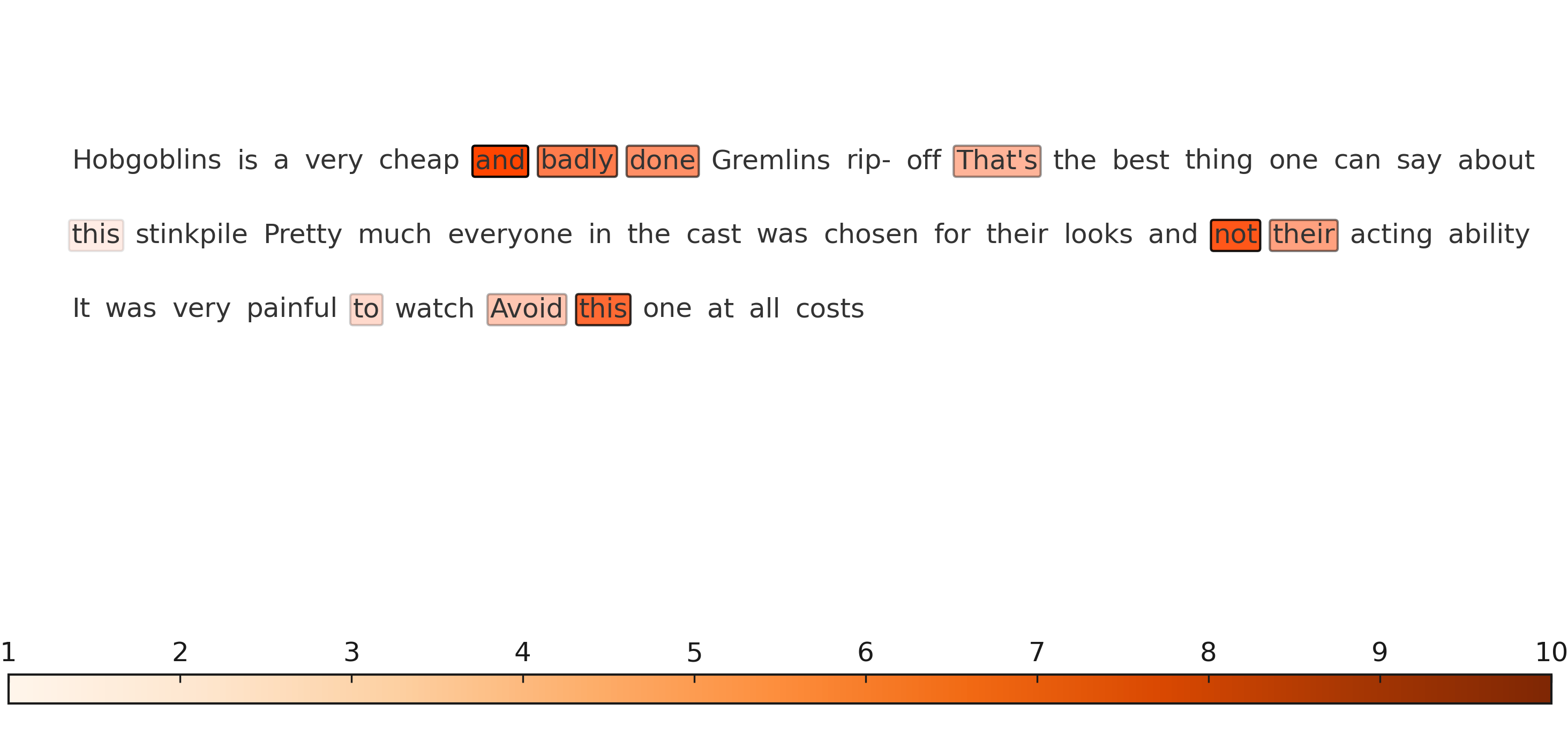}} \\
\hline
\centering\raisebox{12ex}{SIDU-TXT} & \fbox{\includegraphics[width=0.75\linewidth,height=4.6cm]{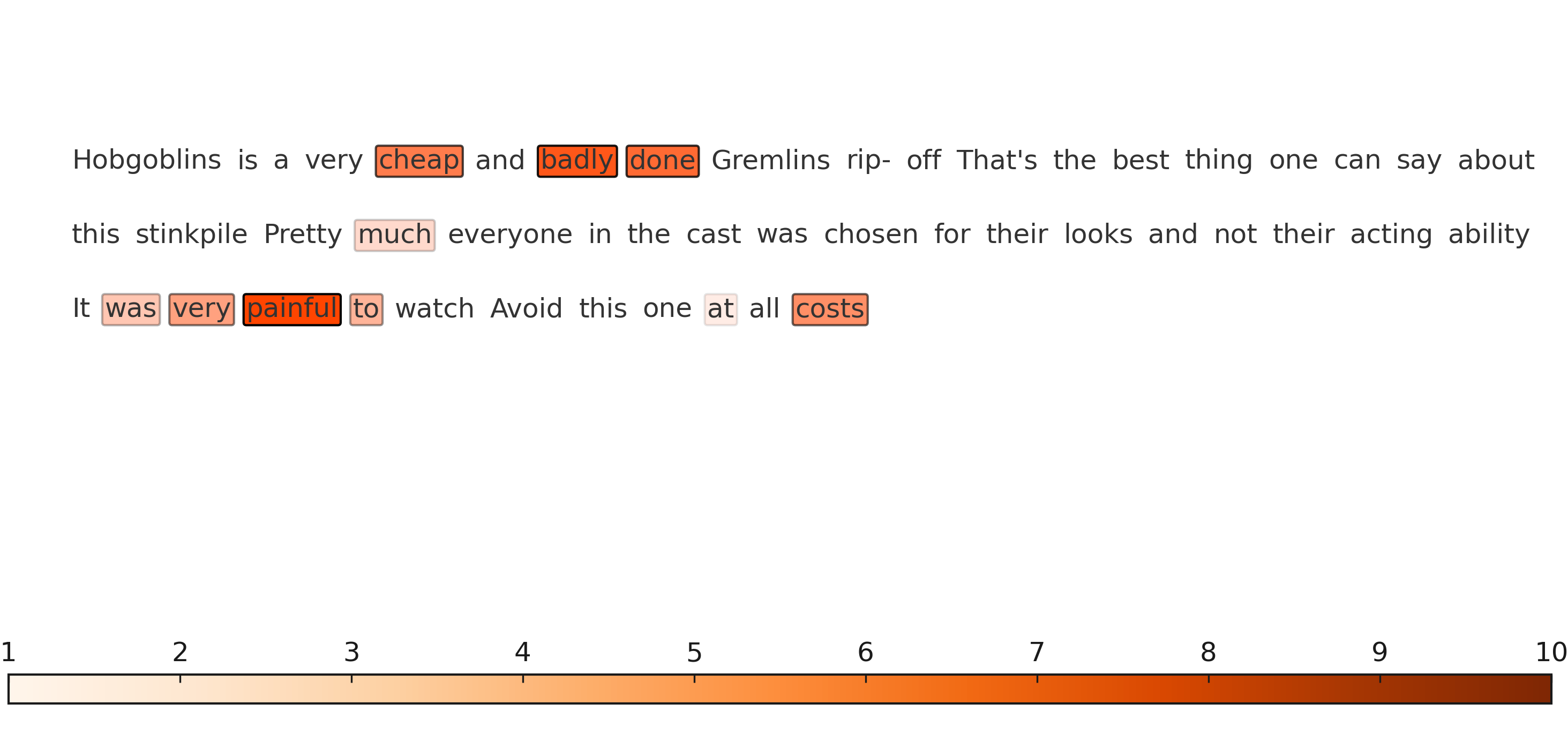}} \\
\hline
\end{tabular}
\label{tab:negative}
\end{table}

In case of LIME, the majority of words are highlighted in shades of orange, suggesting a positive contribution to the sentiment classification. This is in contrast to the review's overall negative sentiment and human annotation. Notably, the word "badly" is highlighted in blue, partially aligning with the human-annotated negative sentiment. The word "avoid" is also marked in blue but with less intensity, suggesting a lesser importance. However, the presence of other blue-highlighted words, which appear to be randomly chosen and irrelevant to the sentiment, adds to the confusion.
 This mixed representation of sentiment by LIME, with a dominance of positive attributions in a negative context, underscores the challenge in ensuring that XAI evaluations provide clear and contextually relevant interpretations for human being. In case of Grad-CAM, the heatmap slightly better aligns with human judgment, spotlighting negative terms such as ''badly done'' and ''avoid this" in orange shades. However, it also applies this highlighting to neutral terms emphasizing same challenge as in the case of LIME. SIDU-TXT's heatmap, on the other hand,  aligns more closely with the words chosen by human annotators, strongly highlighting clear negative words like 'badly done' and 'painful' with the most intense shades of orange. It also picks up additional words that fit the negative sentiment of the review, even if they are not picked up by human annotators (e.g., cheap). Like LIME and Grad-CAM, SIDU-TXT may occasionally highlight irrelevant words, but it marks these less important words with a lower intensity. This shows that SIDU-TXT is is capable of recognizing the relative insignificance of these tokens to a certain extent. A similar analysis and conclusion can be drawn for SIDU-TXT concerning the 'Positive Sentiment' example displayed in Table \ref{tab:po}, where both LIME and Grad-CAM methods exhibit a slight improvement over their performance with the negative example.


\setlength{\tabcolsep}{1pt} 
\renewcommand{\arraystretch}{1.0} 
\begin{table}[H]
\centering
\caption{Comparative heatmaps of the top 10 words from 'Positive sentiment' IMDB dataset reviews. Yellow in Human Annotation denotes equal importance across words. LIME shows negative significance in varying shades of blue and positive in orange, while deeper shades in GRAD-CAM and SIDU-TXT indicate words of greater importance for the XAI methods.}
\begin{tabular}{|c|c|}
\hline
Method & Heatmaps \\
\hline
\centering\raisebox{5.5ex}{Human Annotation} & \fbox{\includegraphics[width=0.75\linewidth]{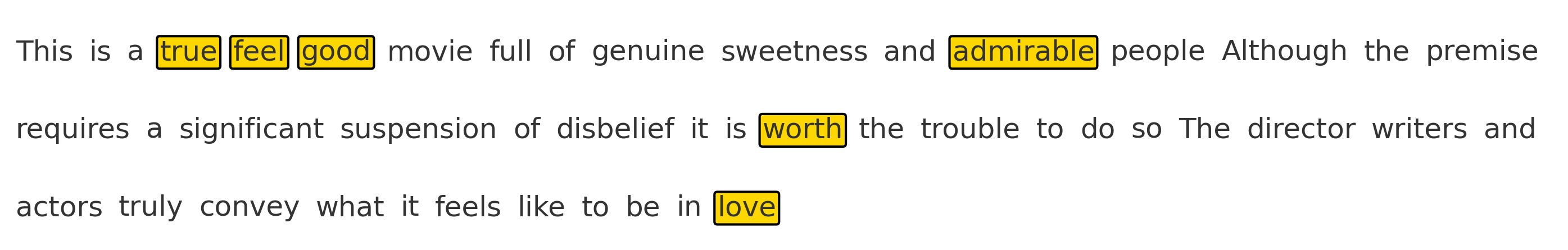}} \\
\hline
\centering\raisebox{12ex}{LIME} & \fbox{\includegraphics[width=0.75\linewidth,height=4.1cm]{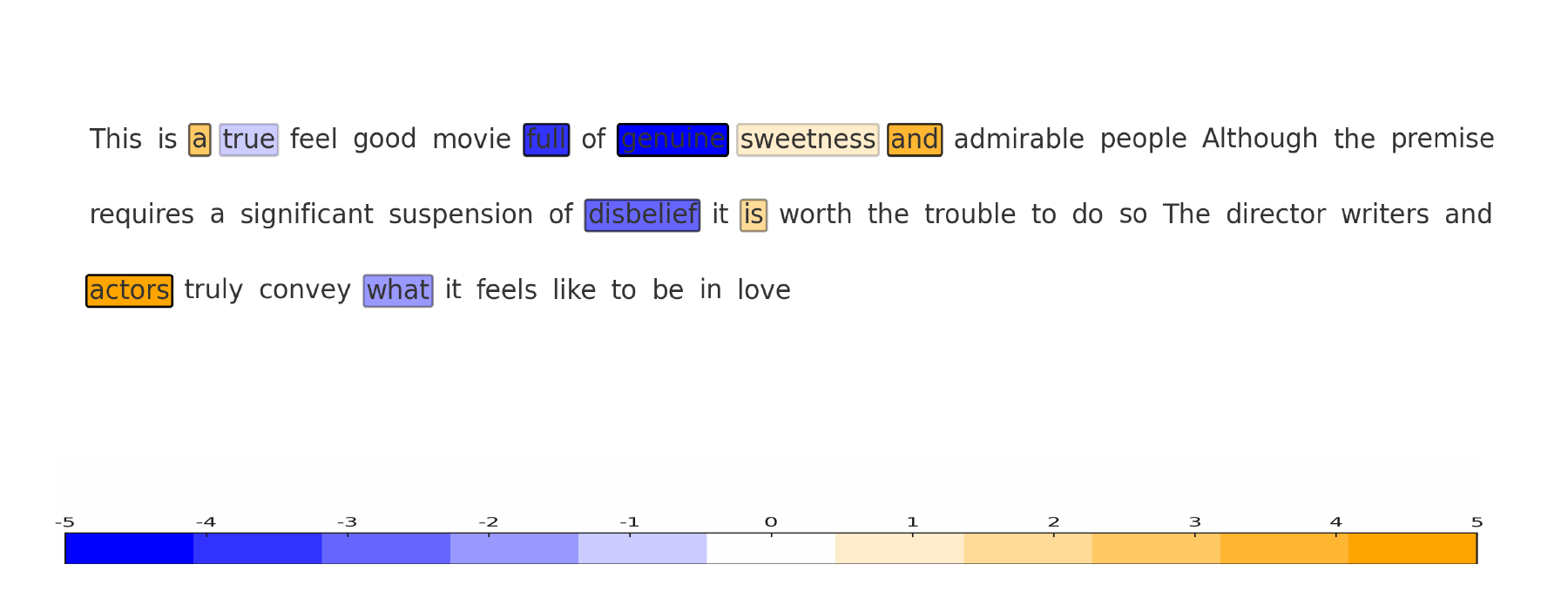}} \\
\hline
\centering\raisebox{12ex}{Grad-CAM}& \fbox{\includegraphics[width=0.75\linewidth,height=4.2cm]{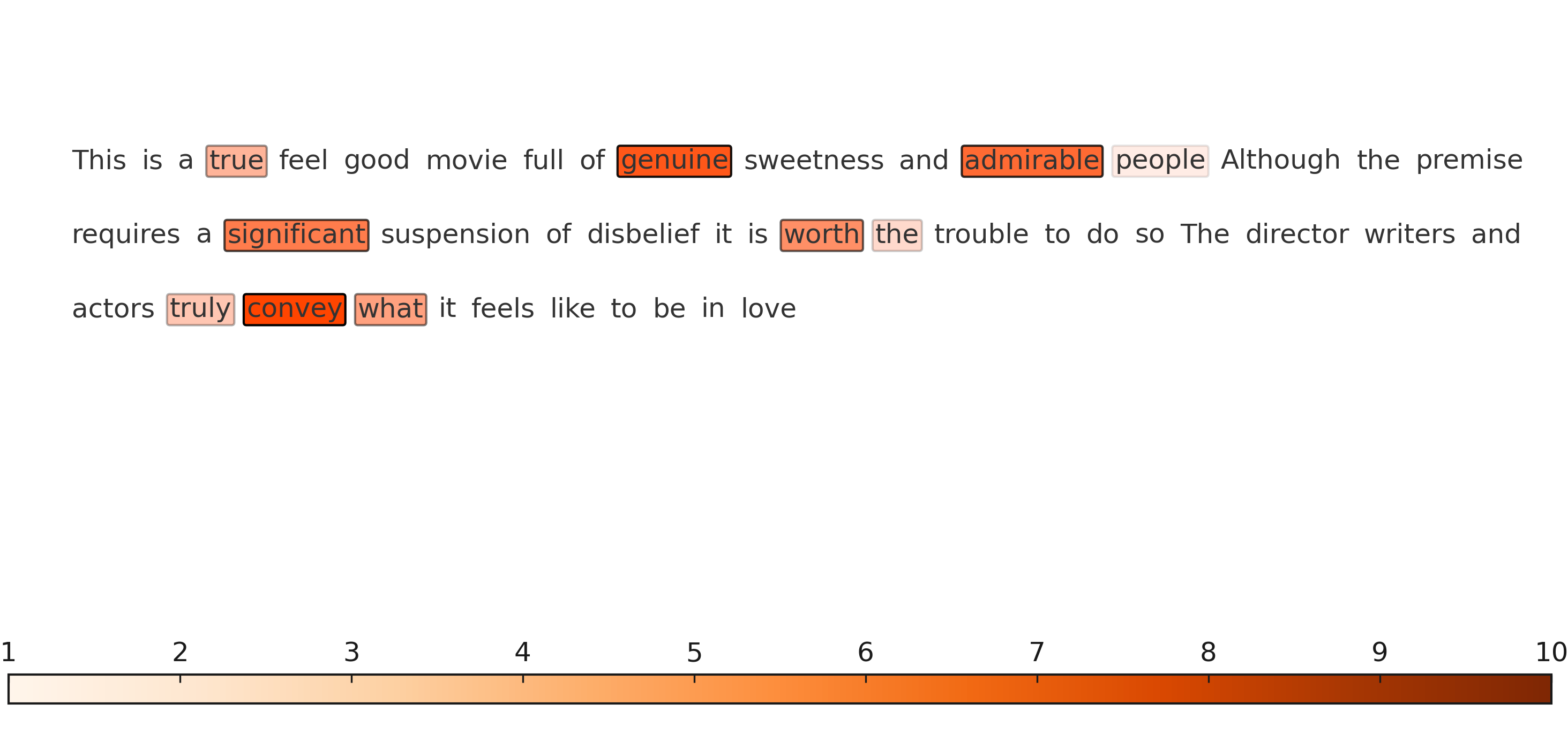}} \\
\hline
\centering\raisebox{12ex}{SIDU-TXT} & \fbox{\includegraphics[width=0.75\linewidth,height=4.3cm]{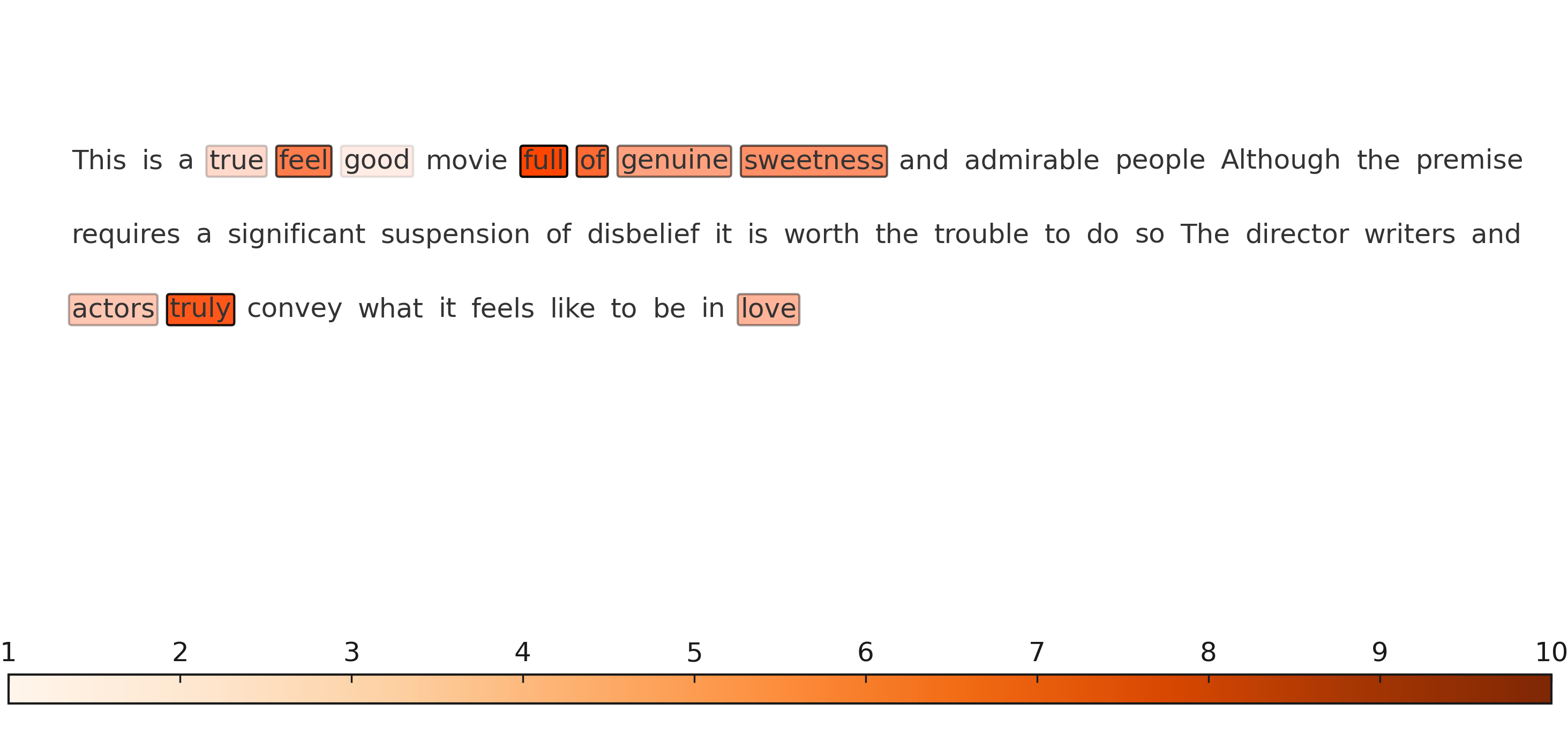}} \\
\hline
\end{tabular}
\label{tab:po}
\end{table}

In summary, while LIME and Grad-CAM provide valuable insights, they occasionally incorporate extraneous words that may lead to misinterpret the sentiment context. This proves once again the inherent challenges in generating explanations that are fully in agreement with human reasoning, a persistent and complex issue in NLP domain.
SIDU-TXT while not being a bulletproof method against similar challenges, appears to have a good grasp of the key terms that convey sentiment, particularly negative sentiment in this case. It further offers a more refined analysis that aligns better with human rationales at the token level. 

In the second experiment, we conducted a quantitative comparison of sentence-level annotations. The primary goal is to evaluate the alignment between XAI methods and human-generated sentence-level rationales. Given that a sentence or phrase often conveys more meaningful context than a single word (token), our focus is on assessing how effectively XAI methods highlight the surrounding context, a group of words, around the most crucial words that contribute meaningful information. The key criterion is whether the XAI method can effectively emphasize sequences of words or phrases, rendering the explanation comprehensible to humans.

For the quantitative evaluation at the sentence level, employing human sentence-level annotations, we treat it as a binary classification problem. We measure the comprehensibility in terms of precision, recall, and F1-score for the human-annotated data samples. Table~\ref{tab:sentence-table}, provides a summary of the mean precision, recall, and F1-score for the XAI methods.


\begin{table}[H]
\centering
\caption{Sentence-level evaluation with Human ground truth}
\begin{tabular}{|l|c|c|c|}
\hline
Method   & Mean Precision $\uparrow$ & Mean Recall$\uparrow$ & Mean F1-SCORE$\uparrow$ \\ \hline
SIDU-TXT & \textbf{0.570}          & \textbf{0.515}       & \textbf{0.504}        \\\hline
GRAD-CAM & 0.399          & 0.484       & 0.383        \\\hline
LIME     & 0.289          & 0.387       & 0.287        \\ \hline
\end{tabular}
\label{tab:sentence-table}
\end{table}

Observing the results, SIDU-TXT methods exhibit high precision, recall, and F1-score, closely followed by Grad-cam. This is evident from the showcased samples in Table 2 and Table 3. The examples illustrate that SIDU can effectively highlight sequences of words adjacent to the most crucial tokens, collectively conveying meaningful information. In contrast, GRAD-CAM and LIME struggle to emphasize sequences of words adjacent to the most crucial tokens. In conclusion, we assert that a robust explanation method should be capable of highlighting sequences of words alongside the most crucial words, ensuring a comprehensible explanation.

\subsubsection{Application-Grounded Evaluation Results}

In our Application-Grounded evaluation, we conduct a qualitative study with legal experts to assess the interpretability of heatmaps generated by the SIDU-TXT, Grad-CAM and LIME XAI methods in the sensitive context of asylum case decision-making. This evaluation is key to understanding the real-world effectiveness of these methods, focusing on the experts' trust in the models' interpretability.

Three researchers, who are experts within both academic and have practitioner experience in asylum law, are presented with heatmaps from these XAI methods for three asylum case files. Each of these case files contains, on average, 5000 words. The experts' assessments focus on the overall explanation quality of each heatmap, particularly how effectively each method explains predictions and highlights legally relevant words and sentences. This involved examining the heatmaps' ability to capture crucial aspects of asylum cases, such as the grounds for claiming asylum, applicant characteristics, and the credibility of evidence. To ensure unbiased evaluations, the experts are not informed about the names of the methods, and the order of presentation is randomized.

SIDU-TXT demonstrated effectiveness in identifying legally relevant information, including details about the asylum claim, travel routes, facts about the motive, and applicant characteristics such as citizenship, religion, and ethnicity. However, it occasionally missed critical details, like documentation of the applicant's experiences, thus affecting its overall completeness in explainability from a traditional legal perspective.

Conversely, Grad-CAM exhibited a more comprehensive approach, with its heatmap effectively highlighting detailed descriptions of facts and evidence, including medical assessments and event chronologies. However, this model also presented challenges, notably in the form of a significant proportion of stop words and false positives, which impacted its precision.

LIME predominantly highlighted stop words, lacking contextual information, although it captured some aspects related to credibility.

In summary, while SIDU-TEXT and Grad-CAM, both belonging to the same class of activation-based XAI methods, succeeded in gaining the expert's trust to varying degrees in highlighting relevant information, both methods exhibited areas of underperformance, indicating the need for further advancing XAI methods for complex real-world applications. The difference in their emphasis on legal reasoning and applicant characteristics highlights an epistemological gap between internal model logic and external validity criteria. This divided judgment, particularly in such intricate and challenging decision-making tasks, experts underscores the need for continuous improvements in model development to bridge this gap.

\section{Conclusion and Future works}\label{sec13}

In this paper, we extend the novel 'Similarity Difference and Uniqueness' (SIDU) XAI method~\cite{muddamsetty2022visual}, originally developed for explaining CNN models in the image domain, to the realm of NLP. Specifically, SIDU-TXT investigates textual predictions through granular,  word-based heatmaps by utilizing feature activation maps of convolution layers in the 1D CNN model. In this adaptation, we acknowledge the importance of striking a balance between the semantic interpretation of contextual text presented in the visual heatmap and the filtering out of noisy, irrelevant non-contextual elements. Consequently, SIDU-TXT focuses on contextually rich features by selecting the top $K$ feature activation masks. This approach provides a deeper understanding of how the proposed XAI model interprets textual data within the decision-making pipeline, ensuring that only the most relevant and informative features are highlighted. 

To evaluate the performance of the proposed SIDU-TXT method, we conducted comprehensive and holistic evaluations focusing on the \textit{faithfulness}, \textit{justifiability}, \textit{comprehensibility}, and \textit{trustworthiness} of explanations using three distinct categories of XAI assessments. They are: Functionally-Grounded (to assess the faithfulness of explanations), Human-Grounded (to assess the justifiability and comprehensibility), and Application-Grounded evaluations (to gain trust from domain experts, examplified in the law domain). For a faithful analysis of the XAI explanations that measure the fidelity of the highlighted features to the underlying decision-making process in the sentiment prediction task, we employ Functionality-Grounded evaluation metrics, specifically insertion and deletion tests. The results demonstrate that SIDU-TXT outperforms both LIME and Grad-CAM in terms of these metrics. Additionally, to assess how closely the generated explanations align with human justifications, we performed a lexical similarity analysis at both the word and sentence levels using the Jaccard similarity coefficient. To understand this agreement between the influential features (words/sentences) identified by laypersons and those highlighted by the XAI methods, we collected human ground-truth sentiment annotations from 50 short movie reviews drawn from the IMDB database. A comparative study of the features highlighted by human annotators and the XAI methods revealed that, in movie sentiment analysis, SIDU-TXT yields Jaccard similarity closer to $1$ at the word level and more closely aligns with human annotations. This close alignment is further qualitatively validated by comparing the top $10$ highlighted words in each XAI method, drawing similar conclusions as in the lexical analysis. Moreover, for comprehensibility in our Human-Grounded evaluation, we perform a quantitative analysis of the contextual similarity between human intuition and the XAI methods at the sentence level. To achieve this, our approach involves assessing similarity via binary classification and computing metrics such as precision, recall, and F1-score. The analysis indicates that SIDU-TXT surpasses Grad-CAM and LIME, effectively suggesting that it can more efficiently exploit and highlight sequences of contextual words adjacent to crucial words that provide comprehensible explanations.
In the final phase of our evaluation framework, the Application-Grounded, we qualitatively assess the power of interpretability in complex real-world applications to gain the trust of experts in the sensitive legal domain. We employ legal experts to qualitatively assess the heatmaps generated by SIDU-TXT, LIME, and Grad-CAM in the asylum case decision-making process. The qualitative assessment reveals that both SIDU-TXT and Grad-CAM outperform LIME in gaining trust from experts, yet each shows strengths and weaknesses in different aspects. However, it is evident that both methods still fall short of the desired level of performance required for an XAI model to be fully utilized in real-life complex NLP tasks. This underscores the need for further research and development in XAI methods to bridge the epistemological gap between current capabilities and the demands of human-centered, high-stakes decision-making environments. These challenges also echo the intricate nature of interpretability in the NLP domain, as discussed in the Related Work section. Specifically, our comprehensive evaluation framework demonstrates SIDU-TXT's superiority in certain contexts, such as sentiment analysis of IMDB movie reviews, but also reveals the difficulty in drawing similar conclusions for more complex tasks like asylum case analyses compared to downstream sentiment analysis tasks. One potential reason for this observed behavior, aside from the inherently challenging semantics of the problem and the high expectations for precision and comprehension in such specific applications, may lie in the foundational model used: for sentiment analysis, we employ a pre-trained Glove embedding, which facilitates better contextual understanding, while in the asylum case study, the absence of such a robust, language and domain-specific foundation model due to limitations in available legal text embeddings, could contribute to the challenges encountered.

Consequently, in future work, we should aim to enhance SIDU-TXT's application in complex NLP tasks by expanding our qualitative evaluations with more case studies with the help of experts and testing the method across various NLP tasks for broader adaptability, and language-specific embeddings as well as developing domain-specific. These steps are crucial for improving interpretability and reliability in sensitive domains, addressing current limitations, and paving the way for more effective XAI methods.

\section*{Declarations}

\begin{itemize}
\item Funding : Research for this publication is principally conducted as part of the Villum Interdisciplinary grant for the project Explainable Artificial Intelligence and Fairness in Asylum Law (XAIfair) and with support from research funded by the Nordic Research Council grant no. 105178 - Nordic Refugee Determination: Advancing Data Science in Migration Law (NoRDASiL), the Volkswagen Foundation Algorithmic Fairness for Asylum Seekers, Refugees (AFAR) and the Danish National Research Foundation Grant no. DNRF169 - Centre of Excellence for Global Mobility Law. It is also a part of the Independent Research Fund Denmark grant for the project Contestable Artificial Intelligence-defining, evaluating and communicating AI contestability in healthcare, law and finance. Additionally, funding is provided by the Responsible AI for Value Creation project (REPAI) funded by the Grundfos Foundation.
\end{itemize}




\end{document}